%% file: paper.tex
\documentclass[11pt,a4paper]{article}
\usepackage[hyperref]{emnlp2018}
\usepackage{times}
\usepackage{latexsym}
\usepackage{CJKutf8}

\usepackage{url}
\usepackage{booktabs}
\usepackage{subcaption}

\definecolor{Bar1}{RGB}{38,120,178} % dark blue
\definecolor{Bar2}{RGB}{253,127,40} % orange

\usepackage{pgfplots}
\pgfplotsset{%
  every axis/.append style={
    font=\sffamily\footnotesize, % sans-serif font
  },
  every non boxed x axis/.append style={ % x axis
    x axis line style=-, % no arrow
  },
  every non boxed y axis/.append style={ % y axis
    y axis line style=-, % no arrow
  },
  every axis legend/.append style={
    at={(0.5,-0.23)},
    anchor=north,
    font=\sffamily\scriptsize,
    draw=none,
    /tikz/every even column/.append style={column sep=3mm}
  },
}

\aclfinalcopy % Uncomment this line for the final submission

\newcommand\ie{i.\,e.}
\newcommand\eg{e.\,g.}
\newcommand\human{\textsc{Human}}
\newcommand\mt{\textsc{MT}}
\newcommand{\Section}[1]{Section~\ref{sec:#1}}

\newcommand{\Figure}[1]{Figure~\ref{fig:#1}}
\newcommand{\Table}[1]{Table~\ref{tab:#1}}

\newcommand{\citeg}[1]{\citeauthor{#1}'s \citeyearpar{#1}} % cite genitive: Martinez et al's (2012)...

\title{Has Machine Translation Achieved Human Parity?\\A Case for Document-level Evaluation}

\author{Samuel Läubli$^1$ \quad Rico Sennrich$^{1,2}$ \quad Martin Volk$^1$ \bigskip\\
  $^1$Institute of Computational Linguistics, University of Zurich\\
  \texttt{\{laeubli,volk\}@cl.uzh.ch} \medskip\\
  $^2$School of Informatics, University of Edinburgh \\
  \texttt{rico.sennrich@ed.ac.uk} \\}

\date{}

\begin{document}
\maketitle

\begin{abstract}
  Recent research suggests that neural machine translation achieves parity with professional human translation on the WMT Chinese--English news translation task.
  We empirically test this claim with alternative evaluation protocols, contrasting the evaluation of single sentences and entire documents.
  In a pairwise ranking experiment, human raters assessing adequacy and fluency show a stronger preference for human over machine translation when evaluating documents as compared to isolated sentences.
  Our findings emphasise the need to shift towards document-level evaluation as machine translation improves to the degree that errors which are hard or impossible to spot at the sentence-level become decisive in discriminating quality of different translation outputs.
\end{abstract}

\section{Introduction}
\label{sec:Introduction}

Neural machine translation \cite{Kalchbrenner2013,Sutskever2014,Bahdanau2014} has
become the de-facto standard in machine translation, outperforming earlier phrase-based approaches in many data settings and shared translation tasks \cite{Luong2015, Sennrich2016, W16-4616}.
Some recent results suggest that neural machine translation ``approaches the accuracy achieved by average bilingual human translators [on some test sets]'' \cite{Wu2016}, or even that its ``translation quality is at human parity when compared to professional human translators'' \cite{Hassan2018}.
Claims of human parity in machine translation are certainly extraordinary, and require extraordinary evidence.\footnote{The term ``parity'' may raise the expectation that there is evidence for equivalence, but the term is used in the definition of ``there [being] no statistical significance between [two outputs] for a test set of candidate translations'' by \citet{Hassan2018}. Still, we consider this finding noteworthy given the strong evaluation setup.}
Laudably, \citet{Hassan2018} have released their data publicly to allow external validation of their claims.
Their claims are further strengthened by the fact that they follow best practices in human machine translation evaluation, using evaluation protocols and tools that are also used at the yearly Conference on Machine Translation (WMT) \cite{Bojar2017},
and take great care in guarding against some confounds such as test set selection and rater inconsistency.

However, the implications of a statistical tie between two machine translation systems in a shared translation task are less severe than that of a statistical tie between a machine translation system and a professional human translator, so we consider the results worthy of further scrutiny.
We perform an independent evaluation of the professional translation and best machine translation system that were found to be of equal quality by \citet{Hassan2018}.
Our main interest lies in the evaluation protocol, and we empirically investigate if the lack of document-level context could explain the inability of human raters to find a quality difference between human and machine translations. We test the following hypothesis:

\begin{quote}
A professional translator who is asked to rank the quality of two candidate translations on the document level will prefer a professional human translation over a machine translation.
\end{quote}

\noindent Note that our hypothesis is slightly different from that tested by \citet{Hassan2018}, which could be phrased as follows:

\begin{quote}
A bilingual crowd worker who is asked to directly assess the quality of candidate translations on the sentence level will prefer a professional human translation over a machine translation.
\end{quote}

% We perform further changes to the evaluation protocol in comparison to \citep{Hassan2018}, believing that they raise the accuracy of human measurements.
% Specifically, our evaluation is based on pairwise ranking rather than direct assessment, and carried out by professional translators rather than crowd workers (\Section{Protocol}).
\noindent As such, our evaluation is not a direct replication of that by \citet{Hassan2018}, and a failure to reproduce their findings does not imply an error on either our or their part.
Rather, we hope to indirectly assess the accuracy of different evaluation protocols.
Our underlying assumption is that professional human translation is still superior to neural machine translation, but that the sensitivity of human raters to these quality differences depends on the evaluation protocol.

\section{Human Evaluation of Machine Translation}

Machine translation is typically evaluated by comparing system outputs to source texts, reference translations, other system outputs, or a combination thereof \citep[for examples, see][]{Bojar2016b}.
The scientific community concentrates on two aspects: adequacy, typically assessed by bilinguals; and target language fluency, typically assessed by monolinguals.
Evaluation protocols have been subject to controversy for decades \citep[\eg,][]{VanSlype1979}, and we identify three aspects with particular relevance to assessing human parity: granularity of measurement (ordinal vs.\ interval scales), raters (experts vs.\ crowd workers), and experimental unit (sentence vs.\ document).

\subsection{Related Work}
\label{sec:RelatedWork}

\paragraph{Granularity of Measurement}

\citet{CallisonBurch2007} show that ranking (\textit{Which of these translations is better?}) leads to better inter-rater agreement than absolute judgement on 5-point Likert scales (\textit{How good is this translation?}) but gives no insight about how much a candidate translation differs from a (presumably perfect) reference.
To this end, \citet{Graham2013} suggest the use of continuous scales for direct assessment of translation quality.
Implemented as a slider between 0~(\textit{Not at all}) and 100~(\textit{Perfectly}), their method yields scores on a 100-point interval scale in practice \citep{Bojar2016,Bojar2017}, with each raters' rating being standardised to increase homogeneity.
\citet{Hassan2018} use source-based direct assessment to avoid bias towards reference translations.
In the shared task evaluation by \citet{Cettolo2017}, raters are shown the source and a candidate text, and asked: \textit{How accurately does the above candidate text convey the semantics of the source text?}
In doing so, they have translations produced by humans and machines rated independently, and parity is assumed if the mean score of the former does not significantly differ from the mean score of the latter.

% Come back to this in 2.2: Conversely, we always show a pair of translations: one produced by a human, one by a machine.

\paragraph{Raters}

To optimise cost, machine translation quality is typically assessed by means of crowdsourcing.
Combined ratings of bilingual crowd workers have been shown to be more reliable than automatic metrics and ``very similar'' to ratings produced by ``experts''\footnote{``Experts'' here are computational linguists who develop MT systems, who may not be expert translators.} \cite{CallisonBurch2009}.
\citet{Graham2017} compare crowdsourced to ``expert'' ratings on machine translations from WMT 2012, concluding that, with proper quality control, ``machine translation systems can indeed be evaluated by the crowd alone.''
However, it is unclear whether this finding carries over to translations produced by NMT systems where, due to increased fluency, errors are more difficult to identify \cite{Castilho17b},
and concurrent work by \citet{Toral2018} highlights the importance of expert translators for MT evaluation.

\paragraph{Experimental Unit}

Machine translation evaluation is predominantly performed on single sentences, presented to raters in random order \citep[\eg,][]{Bojar2017,Cettolo2017}.
There are two main reasons for this. The first is cost:
if raters assess entire documents, obtaining the same number of data points in an evaluation campaign multiplies the cost by the average number of sentences per document. The second is experimental validity.
When comparing systems that produce sentences without considering document-level context, the perceived suprasentential cohesion of a system output is likely due to randomness and thus a confounding factor.
While incorporating document-level context into machine translation systems is an active field of research \cite{Webber2017}, state-of-the-art systems still operate at the level of single sentences \cite{Sennrich2017,NIPS2017_7181,Hassan2018}.
In contrast, human translators can and do take document-level context into account \cite{Krings1986}.
The same holds for raters in evaluation campaigns.
In the discussion of their results, \citet{Wu2016} note that their raters ``[did] not necessarily fully understand each randomly sampled sentence sufficiently'' because it was provided with no context. In such setups, raters cannot reward textual cohesion and coherence.

\subsection{Our Evaluation Protocol}
\label{sec:Protocol}

We conduct a quality evaluation experiment with a 2\,$\times$\,2 mixed factorial design, testing the effect of source text availability (adequacy, fluency) and experimental unit (sentence, document) on ratings by professional translators.

\paragraph{Granularity of Measurement}

We elicit judgements by means of pairwise ranking.
Raters choose the better (with ties allowed) of two translations for each item: one produced by a professional translator (\human), the other by machine translation (\mt).
Since our evaluation includes that of human translation, it is reference-free.
We evaluate in two conditions: adequacy, where raters see source texts and translations (\textit{Which translation expresses the meaning of the source text more adequately?}); and fluency, where raters only see translations (\textit{Which text is better English?}).

\paragraph{Raters}

We recruit professional translators, only considering individuals with at least three years of professional experience and positive client reviews.

\paragraph{Experimental Unit}

To test the effect of context on perceived translation quality, raters evaluate entire documents as well as single sentences in random order (\ie, context is a within-subjects factor).
They are shown both translations (\human{} and \mt) for each unit; the source text is only shown in the adequacy condition.

% Number of articles: 123
% Sentences per article: avg=8.13, stdev=6.46, min=1.00, max=34.00
% ZH source characters per article: avg=1378.04, stdev=1023.04, min=131.00, max=5750.00

\paragraph{Quality Control}

To hedge against random ratings, we convert 5 documents and 16 sentences per set into spam items \cite{Kittur2008}: we render one of the two options nonsensical by shuffling its words randomly, except for 10\,\% at the beginning and end.

\paragraph{Statistical Analysis}
\label{sec:StatisticalAnalysis}

We test for statistically significant preference of \human{} over \mt{} or vice versa by means of two-sided Sign Tests.
Let $a$ be the number of ratings in favour of \mt{}, $b$ the number of ratings in favour of \human{}, and $t$ the number of ties.
We report the number of successes $x$ and the number of trials $n$ for each test, such that $x=b$ and $n=a+b$.\footnote{\citet{EmersonSimon1979} suggest the inclusion of ties such that $x=b+0.5t$ and $n=a+b+t$. This modification has no effect on the significance levels reported in this paper.}

\subsection{Data Collection}

We use the experimental protocol described in the previous section for a quality assessment of Chinese to English translations of news articles. To this end, we randomly sampled 55 documents and 2$\times$120 sentences from the WMT 2017 test set. We only considered the 123 articles (documents) which are native Chinese,\footnote{While it is common practice in machine translation to use the same test set in both translation directions, we consider a direct comparison between human ``translation'' and machine translation hard to interpret if one is in fact the original English text, and the other an automatic translation into English of a human translation into Chinese. In concurrent work, \citet{Toral2018} expand on the confounding effect of evaluating text where the target side is actually the original document.}
containing 8.13 sentences on average. Human and machine translations (\textsc{Reference-HT} as \human, and \textsc{Combo-6} as \mt) were obtained from data released by \citet{Hassan2018}.\footnote{\scriptsize\url{http://aka.ms/Translator-HumanParityData}}

The sampled documents and sentences were rated by professional translators we recruited from ProZ:\footnote{\scriptsize\url{https://www.proz.com}} 4 native in Chinese (2), English (1), or both (1) to rate adequacy, and 4 native in English to rate fluency.
On average, translators had 13.7 years of experience and 8.8 positive client reviews on ProZ, and received US\$ 188.75 for rating 55 documents and 120 sentences.

The averages reported above include an additional translator we recruited when one rater showed poor performance on document-level spam items in the fluency condition, whose judgements we exclude from analysis. We also exclude sentence-level results from 4 raters because there was overlap with the documents they annotated, which means that we cannot rule out that the sentence-level decisions were informed by access to the full document.
To allow for external validation and further experimentation, we make all experimental data publicly available.\footnote{\scriptsize\url{https://github.com/laeubli/parity}}

\section{Results}
\label{sec:Results}

\begin{figure*}
  \begin{subfigure}[b]{0.5\textwidth}
    \centering
    \caption{Adequacy}
    \label{fig:ResultsAdequacy}
    \hspace{-10mm}
    \begin{tikzpicture}
      \begin{axis}[
        ybar, ymin=0, ymax=70,
        width=0.95\textwidth, height=5cm,
        axis x line=bottom,
        axis y line=left,
        ylabel={Preference (\%)},
        y label style={yshift=-1.5mm},
        enlarge x limits=0.5,
        symbolic x coords={MT,Tie,Human},
        xticklabels={\mt{}, Tie, \human{}},
        xtick=data,
        yticklabel={$\mathsf{\pgfmathprintnumber{\tick}}$},
        nodes near coords,
        nodes near coords align={vertical},
        nodes near coords style={font=\sffamily\scriptsize,/pgf/number format/assume math mode},
        legend columns=2,
      ]
      \addplot[color=Bar2,fill=Bar2] coordinates { (MT,50) (Tie,9)  (Human,41) }; % sentence
      \addplot[color=Bar1,fill=Bar1] coordinates { (MT,37) (Tie,11) (Human,52) }; % document
      \legend{Sentence (N=208), Document (N=200)}
      \end{axis}
    \end{tikzpicture}
  \end{subfigure}
  ~
  \begin{subfigure}[b]{0.5\textwidth}
    \centering
    \caption{Fluency}
    \label{fig:ResultsFluency}
    \hspace{-10mm}
    \begin{tikzpicture}
      \begin{axis}[
        ybar, ymin=0, ymax=70,
        width=0.95\textwidth, height=5cm,
        axis x line=bottom,
        axis y line=left,
        ylabel={Preference (\%)},
        y label style={yshift=-1.5mm},
        enlarge x limits=0.5,
        symbolic x coords={MT,Tie,Human},
        xticklabels={\mt{}, Tie, \human{}},
        xtick=data,
        yticklabel={$\mathsf{\pgfmathprintnumber{\tick}}$},
        nodes near coords,
        nodes near coords align={vertical},
        nodes near coords style={font=\sffamily\scriptsize,/pgf/number format/assume math mode},
        legend columns=2,
      ]
      \addplot[color=Bar2,fill=Bar2] coordinates { (MT,32) (Tie,17) (Human,51) }; % sentence
      \addplot[color=Bar1,fill=Bar1] coordinates { (MT,22) (Tie,29) (Human,50) }; % document
      \legend{Sentence (N=208), Document (N=200)}
      \end{axis}
    \end{tikzpicture}
  \end{subfigure}
  \caption{Raters prefer human translation more strongly in entire documents. When evaluating isolated sentences in terms of adequacy, there is no statistically significant difference between \human{} and \mt{}; in all other settings, raters show a statistically significant preference for \human{}.}
  \label{fig:Results}
\end{figure*}
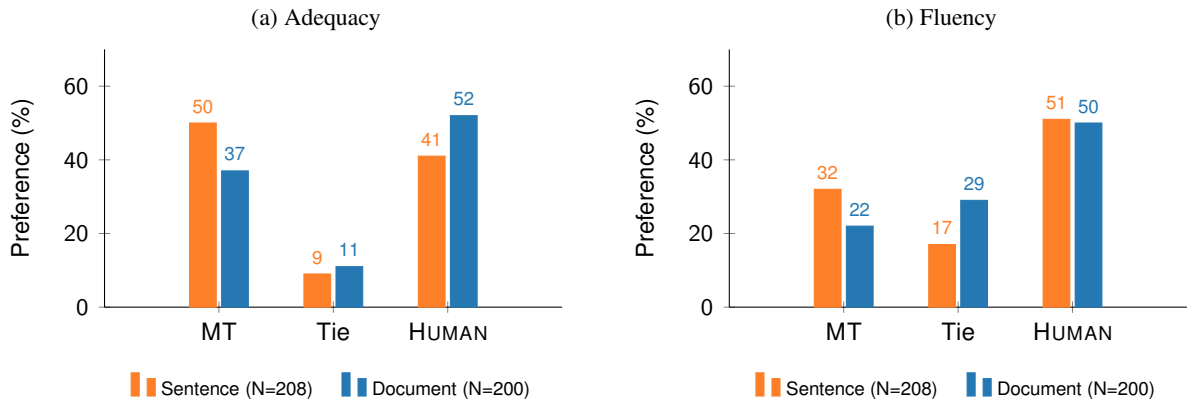

\begingroup
% less space around = and < signs
\setlength{\thinmuskip}{2mu}
\setlength{\medmuskip}{2mu}
\setlength{\thickmuskip}{2mu}

In the adequacy condition, \mt{} and \human{} are not statistically significantly different on the sentence level ($x=86$, $n=189$, $p=.244$).
This is consistent with the results \citet{Hassan2018} obtained with an alternative evaluation protocol (crowdsourcing and direct assessment; see \Section{RelatedWork}).
However, when evaluating entire documents, raters show a statistically significant preference for \human{} ($x=104$, $n=178$, $p<.05$). While the number of ties is similar in sentence- and document-level evaluation, preference for \mt{} drops from 50 to 37\,\% in the latter (\Figure{ResultsAdequacy}).

In the fluency condition, raters prefer \human{} on both the sentence ($x\,=\,106$, $n=172$, $p<.01$) and document level ($x=99$, $n=143$, $p<.001$). In contrast to adequacy, fluency ratings in favour of \human{} are similar in sentence- and document-level evaluation, but raters find more ties with document-level context as preference for \mt{} drops from 32 to 22\,\% (\Figure{ResultsFluency}).

\endgroup

We note that these large effect sizes lead to statistical significance despite modest sample size.
Inter-annotator agreement (Cohen's $\kappa$) ranges from 0.13 to 0.32 (see Appendix for full results and discussion).

\section{Discussion}

Our results emphasise the need for suprasentential context in human evaluation of machine translation.
Starting with \citeg{Hassan2018} finding of no statistically significant difference in translation quality between \human{} and \mt{} for their Chinese--English test set,
we set out to test this result with an alternative evaluation protocol which we expected to strengthen the ability of raters to judge translation quality.
We employed professional translators instead of crowd workers, and pairwise ranking instead of direct assessment, but in a sentence-level evaluation of adequacy, raters still found it hard to discriminate between \human{} and \mt{}:
they did not show a statistically significant preference for either of them.

\begin{CJK*}{UTF8}{gbsn}
Conversely, we observe a tendency to rate \human{} more favourably on the document level than on the sentence level, even within single raters.
Adequacy raters show a statistically significant preference for \human{} when evaluating entire documents.
We hypothesise that document-level evaluation unveils errors such as mistranslation of an ambiguous word, or errors related to textual cohesion and coherence, which remain hard or impossible to spot in a sentence-level evaluation.
For a subset of articles, we elicited both sentence-level and document-level judgements, and inspected articles for which sentence-level judgements were mixed, but where \human{} was strongly preferred in document-level evaluation.
In these articles, we do indeed observe the hypothesised phenomena.
We find an example of lexical coherence in a 6-sentence article about a new app ``微信挪车'', which \human{} consistently translates into ``WeChat Move the Car''.
In \mt{}, we find three different translations in the same article: ``Twitter Move Car'', ``WeChat mobile'', and ``WeChat Move''.
Other observations include the use of more appropriate discourse connectives in \human{}, a more detailed investigation of which we leave to future work.
\end{CJK*}

To our surprise, fluency raters show a stronger preference for \human{} than adequacy raters (\Figure{Results}).
The main strength of neural machine translation in comparison to previous statistical approaches was found to be increased fluency, while adequacy improvements were less clear \citep{Bojar2016,Castilho17}, and we expected a similar pattern in our evaluation.
Does this indicate that adequacy is in fact a strength of \mt{}, not fluency?
We are wary to jump to this conclusion.
An alternative interpretation is that \mt{}, which tends to be more literal than \human{}, is judged more favourably by raters in the bilingual condition, where the majority of raters are native speakers of the source language, because of L1 interference.
We note that the availability of document-level context still has a strong impact in the fluency condition (\Section{Results}).

\section{Conclusions}

In response to recent claims of parity between human and machine translation, we have empirically tested the impact of sentence and document level context on human assessment of machine translation.
Raters showed a markedly stronger preference for human translations when evaluating at the level of documents, as compared to an evaluation of single, isolated sentences.

We believe that our findings have several implications for machine translation research.
Most importantly, if we accept our interpretation that human translation is indeed of higher quality in the dataset we tested, this points to a failure of current best practices in machine translation evaluation.
As machine translation quality improves, translations will become harder to discriminate in terms of quality, and it may be time to shift towards document-level evaluation, which gives raters more context to understand the original text and its translation,
and also exposes translation errors related to discourse phenomena which remain invisible in a sentence-level evaluation.

Our evaluation protocol was designed with the aim of providing maximal validity, which is why we chose to use professional translators and pairwise ranking.
% Other important factors in evaluation are cost and scalability to the evaluation of many systems, which are strong reasons for the use of crowdsourcing and direct assessment.
For future work, it would be of high practical relevance to test whether we can also elicit accurate quality judgements on the document-level via crowdsourcing and direct assessment, or via alternative evaluation protocols.
The data released by \citet{Hassan2018} could serve as a test bed to this end.

One reason why document-level evaluation widens the quality gap between machine translation and human translation is that the machine translation system we tested still operates on the sentence level, ignoring wider context.
It will be interesting to explore to what extent existing and future techniques for document-level machine translation can narrow this gap.
We expect that this will require further efforts in creating document-level training data, designing appropriate models, and supporting research with discourse-aware automatic metrics.

\section*{Acknowledgements}

We thank Xin Sennrich for her help with the \mbox{analysis} of translation errors. We also thank Antonio Toral and the anonymous reviewers for their helpful comments.

\bibliography{references}
\bibliographystyle{acl_natbib_nourl}

\newpage

\input{appendix}

\end{document}

%% file: appendix.tex
\appendix

\section{Further Statistical Analysis}
\label{sec:FurtherStatisticalAnalysis}

\medskip

Table~\ref{tab:Results} shows detailed results, including those of individual raters, for all four experimental conditions.
Raters choose between three labels for each item: \mt{} is better than \human{} ($a$), \human{} is better than \mt{} ($b$), or tie ($t$). \Table{IAA} lists inter-rater agreement.
Besides percent agreement (same label), we calculate Cohen's kappa coefficient

\medskip

\begin{equation}
  \kappa = \frac{P(A) - P(E)}{1 - P(E)}\,,
\end{equation}

\medskip

\noindent where $P(A)$ is the proportion of times that two raters agree, and $P(E)$ the likelihood of agreement by chance.
We calculate Cohen's kappa, and specifically $P(E)$, as in WMT \citep[Section~3.3]{Bojar2016}, on the basis of all pairwise ratings across all raters.

In pairwise rankings of machine translation outputs, $\kappa$ coefficients typically centre around 0.3 \citep{Bojar2016}.
We observe lower inter-rater agreement in three out of four conditions, and attribute this to two reasons.
Firstly, the quality of the machine translations produced by \citet{Hassan2018} is high, making it difficult to discriminate from professional translation particularly at the sentence level.
Secondly, we do not provide guidelines detailing error severity and thus assume that raters have differing interpretations of what constitutes a ``better'' or ``worse'' translation.
Confusion matrices in \Table{CF} indicate that raters handle ties very differently: in document-level adequacy, for example, rater E assigns no ties at all, while rater F rates 15 out of 50 items as ties (\Table{CF:g}).
The assignment of ties is more uniform in documents assessed for fluency  (Tables~\ref{tab:Results},~\ref{tab:CF:a}--\ref{tab:CF:f}), leading to higher $\kappa$ in this condition (\Table{IAA}).

Despite low inter-annotator agreement, the quality control we apply shows that raters assess items carefully: they only miss 1 out of 40 and 5 out of 128 spam items in the document- and sentence-level conditions overall, respectively, a very low number compared to crowdsourced work \cite{Kittur2008}.
All of these misses are ties (\ie, not marking spam items as ``better'', but rather equally bad as their counterpart), and 5 out of 9 raters (A, B1, B2, D, F) do not miss a single spam item.

\newpage

\noindent A common procedure in situations where inter-rater agreement is low is to aggregate ratings of different annotators \citep{Graham2017}. As shown in \Table{MajorityVoting}, majority voting leads to clearer discrimination between \mt{} and \human{} in all conditions, except for sentence-level adequacy.

\bigskip

\begin{table}[h!]
\small
\begin{tabular}{lrrrrrr}
\toprule \\ [-2ex]
& \multicolumn{3}{c}{Document} & \multicolumn{3}{c}{Sentence}\\
\cmidrule(lr){2-4} \cmidrule(lr){5-7}
Rater & MT & Tie & \hspace{-2ex}Human & MT & Tie & \hspace{-2ex}Human \\ [0.5ex]
\hline \\ [-1.5ex]
Fluency \\ [0.5ex]
\quad A & 13 & 8 & 29 & 30 & 32 & 42 \\
\quad B1 & & &        & 36 &  4 & 64 \\
\quad B2 & 8 & 18 & 24 \\
\quad C & 12 & 14 & 24 & \textcolor{gray}{40} & \textcolor{gray}{14} & \textcolor{gray}{50} \\
\quad D & 11 & 17 & 22 & \textcolor{gray}{32} & \textcolor{gray}{30} & \textcolor{gray}{42} \\ [1.0ex]
\quad total & 44 & 57 & 99   & 66 & 36 & 106 \\ [0.5ex]
% \quad Majority & 24 & 10 & 66 \\ [1.0ex]
\hline \\ [-1.5ex]
Adequacy \\ [0.5ex]
\quad E & 26 & 0  & 24 & 59 &  3 & 42\\
\quad F & 10 & 15 & 25 & 44 & 16 & 44 \\
\quad G & 18 &  4 & 28 & \textcolor{gray}{38} & \textcolor{gray}{23} & \textcolor{gray}{43} \\
\quad H & 20 &  3 & 27 & \textcolor{gray}{38} & \textcolor{gray}{11} & \textcolor{gray}{55} \\ [1.0ex]
\quad total & 74 & 22 & 104   & 103 & 19 & 86 \\
% \quad Majority & 32 & 18 & 50 \\ [1.0ex]
\bottomrule \\ [-3.5ex]
\end{tabular}
\caption{Ratings by rater and condition.
Greyed-out fields indicate that raters had access to full documents for which we elicited sentence-level judgements; these are not considered for total results.}
\label{tab:Results}
\end{table}

\begin{table}[h!]
\small
\begin{tabular}{lcccccc}
\toprule \\ [-1.0ex]
& \multicolumn{3}{c}{Document} & \multicolumn{3}{c}{Sentence}\\
\cmidrule(lr){2-4} \cmidrule(lr){5-7}
Aggregation & MT & Tie & \hspace{-2ex}Human & MT & Tie & \hspace{-2ex}Human \\ [0.8ex]
\hline \\ [-1.0ex]
Fluency \\ [0.5ex]
\quad Average  & 22 & 29 & 50   & 32 & 17 & 51 \\
\quad Majority & 24 & 10 & 66   & 26 & 23 & 51 \\ [1.0ex]
\hline \\ [-1.0ex]
Adequacy \\ [0.5ex]
\quad Average  & 37 & 11 & 52   & 50 &  9 & 41 \\
\quad Majority & 32 & 18 & 50   & 38 & 32 & 31 \\ [1.0ex]
\bottomrule
\end{tabular}
\caption{Aggregation of ratings (\%).}
\label{tab:MajorityVoting}
\end{table}

\begin{table}[h!]
\small
\centering
  \begin{tabular}{lrr}
    \toprule \\ [-1.5ex]
      & Document  & Sentence  \\ [0.5ex]
      \hline \\ [-1.5ex]
      Fluency             &           &           \\
      \quad Same label        & 55\,\%    & 45\,\%    \\
      \quad Cohen's $\kappa$  & 0.32      & 0.13      \\ [0.5ex]
      \hline \\ [-1.5ex]
      Adequacy               &           &           \\
      \quad Same label        & 49\,\%    & 50\,\%    \\
      \quad Cohen's $\kappa$  & 0.13      & 0.14      \\
    \bottomrule
  \end{tabular}
  \caption{Inter-rater agreement.}
  \label{tab:IAA}
\end{table}

\input{confusion_matrices}

%% file: confusion_matrices.tex
\begin{table*}
  \centering

  % line 1
  \begin{subtable}[b]{0.3\textwidth}
    \small
    \renewcommand{\arraystretch}{1.5}
    \begin{tabular}{rr|rrr|}
      \multicolumn{2}{c}{} & \multicolumn{3}{c}{B2} \\
          & \multicolumn{1}{c}{} & $a$ & $t$ & \multicolumn{1}{c}{$b$} \\
            \cmidrule{3-5}   \\ [-4.5ex]
          & $a$ & 5 & 4 & 4  \\
      A   & $t$ & 1 & 5 & 2  \\
          & $b$ & 2 & 9 & 18 \\ [-0.5ex]
            \cmidrule{3-5}
    \end{tabular}
    \caption{fluency, document, N=50}
    \label{tab:CF:a}
  \end{subtable}
  ~
  \begin{subtable}[b]{0.3\textwidth}
    \small
    \renewcommand{\arraystretch}{1.5}
    \begin{tabular}{rr|rrr|}
      \multicolumn{2}{c}{} & \multicolumn{3}{c}{C} \\
          & \multicolumn{1}{c}{} & $a$ & $t$ & \multicolumn{1}{c}{$b$} \\
            \cmidrule{3-5}   \\ [-4.5ex]
          & $a$ & 7 & 2 & 4  \\
      A   & $t$ & 2 & 4 & 2  \\
          & $b$ & 3 & 8 & 18 \\ [-0.5ex]
            \cmidrule{3-5}
    \end{tabular}
    \caption{fluency, document, N=50}
    \label{tab:CF:b}
  \end{subtable}
  ~
  \begin{subtable}[b]{0.3\textwidth}
    \small
    \renewcommand{\arraystretch}{1.5}
    \begin{tabular}{rr|rrr|}
      \multicolumn{2}{c}{} & \multicolumn{3}{c}{D} \\
          & \multicolumn{1}{c}{} & $a$ & $t$ & \multicolumn{1}{c}{$b$} \\
            \cmidrule{3-5}   \\ [-4.5ex]
          & $a$ & 6 & 3 & 4  \\
      A   & $t$ & 2 & 6 & 0  \\
          & $b$ & 3 & 8 & 18 \\ [-0.5ex]
            \cmidrule{3-5}
    \end{tabular}
    \caption{fluency, document, N=50}
    \label{tab:CF:c}
  \end{subtable}

  \vspace{4ex}

  % line 2
  \begin{subtable}[b]{0.3\textwidth}
    \small
    \renewcommand{\arraystretch}{1.5}
    \begin{tabular}{rr|rrr|}
      \multicolumn{2}{c}{} & \multicolumn{3}{c}{C} \\
          & \multicolumn{1}{c}{} & $a$ & $t$ & \multicolumn{1}{c}{$b$} \\
            \cmidrule{3-5}   \\ [-4.5ex]
          & $a$ & 5 & 1 & 2  \\
      B2  & $t$ & 4 & 5 & 9  \\
          & $b$ & 3 & 8 & 13 \\ [-0.5ex]
            \cmidrule{3-5}
    \end{tabular}
    \caption{fluency, document, N=50}
    \label{tab:CF:d}
  \end{subtable}
  ~
  \begin{subtable}[b]{0.3\textwidth}
    \small
    \renewcommand{\arraystretch}{1.5}
    \begin{tabular}{rr|rrr|}
      \multicolumn{2}{c}{} & \multicolumn{3}{c}{D} \\
          & \multicolumn{1}{c}{} & $a$ & $t$ & \multicolumn{1}{c}{$b$} \\
            \cmidrule{3-5}   \\ [-4.5ex]
          & $a$ & 6 & 1 & 1  \\
      B2  & $t$ & 3 & 7 & 8  \\
          & $b$ & 2 & 9 & 13 \\ [-0.5ex]
            \cmidrule{3-5}
    \end{tabular}
    \caption{fluency, document, N=50}
    \label{tab:CF:e}
  \end{subtable}
  ~
  \begin{subtable}[b]{0.3\textwidth}
    \small
    \renewcommand{\arraystretch}{1.5}
    \begin{tabular}{rr|rrr|}
      \multicolumn{2}{c}{} & \multicolumn{3}{c}{D} \\
          & \multicolumn{1}{c}{} & $a$ & $t$ & \multicolumn{1}{c}{$b$} \\
            \cmidrule{3-5}   \\ [-4.5ex]
          & $a$ & 7 & 3 & 2  \\
      C & $t$ & 1 & 7 & 6  \\
          & $b$ & 3 & 7 & 14 \\ [-0.5ex]
            \cmidrule{3-5}
    \end{tabular}
    \caption{fluency, document, N=50}
    \label{tab:CF:f}
  \end{subtable}

  \vspace{4ex}

  % line 3
  \begin{subtable}[b]{0.3\textwidth}
    \small
    \renewcommand{\arraystretch}{1.5}
    \begin{tabular}{rr|rrr|}
      \multicolumn{2}{c}{} & \multicolumn{3}{c}{F} \\
          & \multicolumn{1}{c}{} & $a$ & $t$ & \multicolumn{1}{c}{$b$} \\
            \cmidrule{3-5}   \\ [-4.5ex]
          & $a$ & 4 & 9 & 13  \\
      E   & $t$ & 0 & 0 & 0  \\
          & $b$ & 6 & 6 & 12 \\ [-0.5ex]
            \cmidrule{3-5}
    \end{tabular}
    \caption{adequacy, document, N=50}
    \label{tab:CF:g}
  \end{subtable}
  ~
  \begin{subtable}[b]{0.3\textwidth}
    \small
    \renewcommand{\arraystretch}{1.5}
    \begin{tabular}{rr|rrr|}
      \multicolumn{2}{c}{} & \multicolumn{3}{c}{G} \\
          & \multicolumn{1}{c}{} & $a$ & $t$ & \multicolumn{1}{c}{$b$} \\
            \cmidrule{3-5}   \\ [-4.5ex]
          & $a$ & 9 & 4 & 13  \\
      E   & $t$ & 0 & 0 & 0  \\
          & $b$ & 9 & 0 & 15 \\ [-0.5ex]
            \cmidrule{3-5}
    \end{tabular}
    \caption{adequacy, document, N=50}
    \label{tab:CF:h}
  \end{subtable}
  ~
  \begin{subtable}[b]{0.3\textwidth}
    \small
    \renewcommand{\arraystretch}{1.5}
    \begin{tabular}{rr|rrr|}
      \multicolumn{2}{c}{} & \multicolumn{3}{c}{H} \\
          & \multicolumn{1}{c}{} & $a$ & $t$ & \multicolumn{1}{c}{$b$} \\
            \cmidrule{3-5}   \\ [-4.5ex]
          & $a$ & 11 & 1 & 14  \\
      E   & $t$ & 0 & 0 & 0  \\
          & $b$ & 9 & 2 & 13 \\ [-0.5ex]
            \cmidrule{3-5}
    \end{tabular}
    \caption{adequacy, document, N=50}
    \label{tab:CF:i}
  \end{subtable}

  \vspace{4ex}

  % line 4
  \begin{subtable}[b]{0.3\textwidth}
    \small
    \renewcommand{\arraystretch}{1.5}
    \begin{tabular}{rr|rrr|}
      \multicolumn{2}{c}{} & \multicolumn{3}{c}{G} \\
          & \multicolumn{1}{c}{} & $a$ & $t$ & \multicolumn{1}{c}{$b$} \\
            \cmidrule{3-5}   \\ [-4.5ex]
          & $a$ & 7 & 1 & 2  \\
      F   & $t$ & 7 & 1 & 7  \\
          & $b$ & 4 & 2 & 19 \\ [-0.5ex]
            \cmidrule{3-5}
    \end{tabular}
    \caption{adequacy, document, N=50}
    \label{tab:CF:j}
  \end{subtable}
  ~
  \begin{subtable}[b]{0.3\textwidth}
    \small
    \renewcommand{\arraystretch}{1.5}
    \begin{tabular}{rr|rrr|}
      \multicolumn{2}{c}{} & \multicolumn{3}{c}{H} \\
          & \multicolumn{1}{c}{} & $a$ & $t$ & \multicolumn{1}{c}{$b$} \\
            \cmidrule{3-5}   \\ [-4.5ex]
          & $a$ & 6 & 1 & 3  \\
      F   & $t$ & 8 & 0 & 7  \\
          & $b$ & 6 & 2 & 17 \\ [-0.5ex]
            \cmidrule{3-5}
    \end{tabular}
    \caption{adequacy, document, N=50}
    \label{tab:CF:k}
  \end{subtable}
  ~
  \begin{subtable}[b]{0.3\textwidth}
    \small
    \renewcommand{\arraystretch}{1.5}
    \begin{tabular}{rr|rrr|}
      \multicolumn{2}{c}{} & \multicolumn{3}{c}{H} \\
          & \multicolumn{1}{c}{} & $a$ & $t$ & \multicolumn{1}{c}{$b$} \\
            \cmidrule{3-5}   \\ [-4.5ex]
          & $a$ & 11 & 2 & 5  \\
      G   & $t$ & 1 & 1 & 2  \\
          & $b$ & 8 & 0 & 20 \\ [-0.5ex]
            \cmidrule{3-5}
    \end{tabular}
    \caption{adequacy, document, N=50}
    \label{tab:CF:l}
  \end{subtable}

  \vspace{4ex}

  % line 5
  \begin{subtable}[b]{0.3\textwidth}
    \small
    \renewcommand{\arraystretch}{1.5}
    \begin{tabular}{rr|rrr|}
      \multicolumn{2}{c}{} & \multicolumn{3}{c}{B1} \\
          & \multicolumn{1}{c}{} & $a$ & $t$ & \multicolumn{1}{c}{$b$} \\
            \cmidrule{3-5}   \\ [-4.5ex]
          & $a$ & 16 & 1 & 13  \\
      A   & $t$ & 10 & 1 & 21  \\
          & $b$ & 10 & 2 & 30 \\ [-0.5ex]
            \cmidrule{3-5}
    \end{tabular}
    \caption{fluency, sentence, N=104}
    \label{tab:CF:m}
  \end{subtable}
  ~
  \begin{subtable}[b]{0.3\textwidth}
    \small
    \renewcommand{\arraystretch}{1.5}
    \begin{tabular}{rr|rrr|}
      \multicolumn{2}{c}{} & \multicolumn{3}{c}{F} \\
          & \multicolumn{1}{c}{} & $a$ & $t$ & \multicolumn{1}{c}{$b$} \\
            \cmidrule{3-5}   \\ [-4.5ex]
          & $a$ & 31 & 6 & 22  \\
      E   & $t$ & 2 & 0 & 1  \\
          & $b$ & 11 & 10 & 21 \\ [-0.5ex]
            \cmidrule{3-5}
    \end{tabular}
    \caption{adequacy, sentence, N=104}
    \label{tab:CF:n}
  \end{subtable}

  \vspace{4ex}

  \caption{Confusion matrices: \mt{} is better than \human{} ($a$), \human{} is better than \mt{} ($b$), or tie ($t$). Participant IDs (A--H) are the same as in Table~\ref{tab:Results}.}
  \label{tab:CF}
\end{table*}